\documentclass[conference]{IEEEtran}
\IEEEoverridecommandlockouts
\usepackage{cite}
\usepackage{amsmath,amssymb,amsfonts}
\usepackage{algorithmic}
\usepackage{graphicx}
\usepackage{textcomp}
\usepackage{xcolor}
\usepackage{caption}
\newcommand\correspondingauthor{\thanks{Corresponding: Guodong Long \{guodong.long@uts.edu.au\}}}
\usepackage{subcaption}
\def\BibTeX{{\rm B\kern-.05em{\sc i\kern-.025em b}\kern-.08em
    T\kern-.1667em\lower.7ex\hbox{E}\kern-.125emX}}

\usepackage{amsmath,amsfonts,bm}

\def\eqref#1{equation~\ref{#1}}

\def\1{\bm{1}}

\def\vb{{\bm{b}}}

\def\vh{{\bm{h}}}

\def\vw{{\bm{w}}}

\def\mE{{\bm{E}}}

\def\mG{{\bm{G}}}

\def\mK{{\bm{K}}}

\def\mM{{\bm{M}}}

\def\mP{{\bm{P}}}
\def\mQ{{\bm{Q}}}

\def\mV{{\bm{V}}}
\def\mW{{\bm{W}}}

\DeclareMathAlphabet{\mathsfit}{\encodingdefault}{\sfdefault}{m}{sl}
\SetMathAlphabet{\mathsfit}{bold}{\encodingdefault}{\sfdefault}{bx}{n}
\newcommand{\tens}[1]{\bm{\mathsfit{#1}}}

\def\tE{{\tens{E}}}

\def\gG{{\mathcal{G}}}

\def\sC{{\mathbb{C}}}

\def\sN{{\mathbb{N}}}

\newcommand{\parents}{Pa} 

\begin{document}

\title{Sequential Diagnosis Prediction with Transformer and Ontological Representation
}

\author{
\IEEEauthorblockN{
    Xueping Peng\IEEEauthorrefmark{1},
    Guodong Long\IEEEauthorrefmark{1}\correspondingauthor,
    Tao Shen\IEEEauthorrefmark{1},
    Sen Wang\IEEEauthorrefmark{2},
    Jing Jiang\IEEEauthorrefmark{1}
}
\IEEEauthorblockA{
    \IEEEauthorrefmark{1} Australian Artificial Intelligence Institute, FEIT, University of Technology Sydney, Australia \\ 
    \IEEEauthorrefmark{2} School of Information Technology and Electrical Engineering, The University of Queensland, Australia \\
    Email: \{xueping.peng, guodong.long, tao.shen, jing.jiang\}@uts.edu.au, sen.wang@uq.edu.au}
}


\maketitle

\begin{abstract}
Sequential diagnosis prediction on the Electronic Health Record (EHR) has been proven crucial for predictive analytics in the medical domain. 
EHR data, sequential records of a patient's interactions with healthcare systems, has numerous inherent characteristics of temporality, irregularity and data insufficiency.
Some recent works train healthcare predictive models by making use of sequential information in EHR data, but they are vulnerable to irregular, temporal EHR data with the states of admission/discharge from hospital, and insufficient data.
To mitigate this, we propose an end-to-end robust transformer-based model called SETOR, which exploits neural ordinary differential equation to handle both irregular intervals between a patient's visits with admitted timestamps and length of stay in each visit, to alleviate the limitation of insufficient data by integrating medical ontology, and to capture the dependencies between the patient's visits by employing multi-layer transformer blocks.
Experiments conducted on two real-world healthcare datasets show that, our sequential diagnoses prediction model SETOR not only achieves better predictive results than previous state-of-the-art approaches, irrespective of sufficient or insufficient training data, but also derives more interpretable embeddings of medical codes. The experimental codes are available at the GitHub repository\footnote{Github repository: https://github.com/Xueping/SETOR}. 
\end{abstract}


\section{Introduction}
\begin{figure}[t]
  \includegraphics[width=0.48\textwidth]{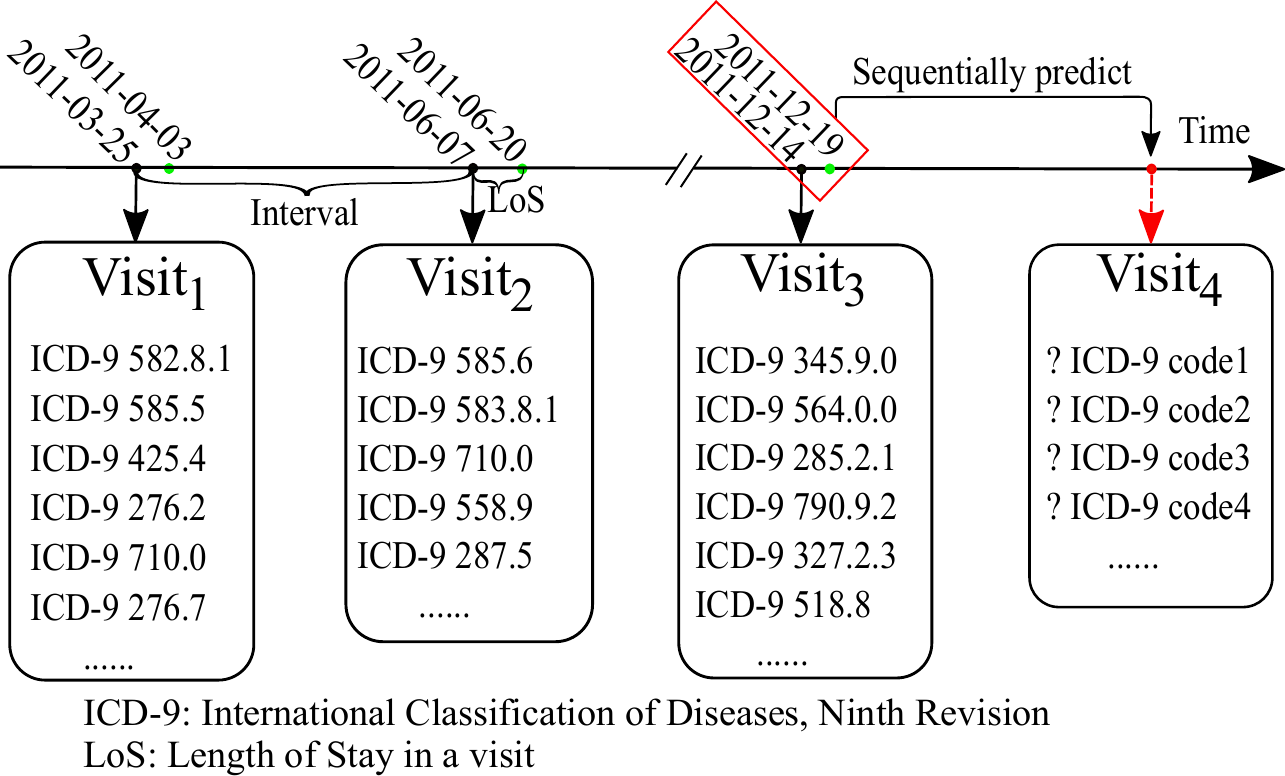}
  \caption{An EHR segment and prediction task. The black and green dots indicate the admission and discharge dates, respectively.}
  \label{fig:example}
\end{figure}

With the rapid growth of the utilization of healthcare information systems during the last few decades, huge volumes of electronic health records (EHR) have been accumulated. 
The patient EHR data typically consists of a sequence of visit records with irregular admitted intervals, and each visit consists of admission and discharge timestamps and a set of clinical events, such as diagnoses, procedures, medications, etc.~\cite{Shickel_2018,song2019medical}. Fig.~\ref{fig:example} shows an EHR segment of a patient, which is referred to as the patient journey in the paper. 
Analyzing the EHR data to benefit the care for a large number of patients has been attracting tremendous attentions from both academia and industry. One of the numerous  analytical tasks is to predict the future diagnoses~\cite{Choi_Bahadori_2017_gram,Ma2018-gu_kame,song2019medical,peng2020self,chen2018dynamic} based on a patient’s historical EHR data. For example, \cite{Choi_Bahadori_2017_gram} and~\cite{Ma2018-gu_kame} employ recurrent neural networks (RNN) to integrate medical ontology for capturing temporal visits and predicting sequential diagnoses. Ref. \cite{song2019medical} predict future diagnoses by utilizing co-occurrence statistics and multiple ontological representations via attention mechanism on EHR data. Although the existing methods have achieved promising results, they are still challenged by the following two limitations.

One of the major challenges is how to effectively model such irregular and temporal EHR data with admission and discharge states from hospital. 
Some recent works~\cite{Choi_2016_med2vec,peng2020self,zhang2019interpretable,peng2020bitenet,Ma2017-gs_Dipole,Choi_Bahadori_2017_gram,choi2016retain,Ma2018-gu_kame,jha2018interpretable} directly adapt text representation learning algorithms~\cite{Mikolov_2013_b} to the sequence-formatted EHR data. 
For example, Med2Vec~\cite{Choi_2016_med2vec} learns a vector representation for each medical concept from the co-occurrence information without considering the temporal sequential nature of the EHR data. 
Further, considering both long-term dependency and sequential information, recurrent neural networks~\cite{Ma2017-gs_Dipole,Choi_Bahadori_2017_gram,choi2016retain,Ma2018-gu_kame,qiao2018pairwise}, including LSTM~\cite{hochreiter1997long} and GRU~\cite{cho2014learning}, are used to learn the contextualized representation of EHR data. 
However, despite the similarity between EHR data and natural language text, one major difference is that EHR data inherently includes the timestamp property. Namely, beyond dependency, there is time interval between each pair of visits.
For example, as shown in Fig.~\ref{fig:example}, the interval between Visit1 and Visit2 is shorter than that between Visit2 and Visit3, indicating that the dependency between Visit1 and Visit2 may be stronger than that later one. Due to irregular visits of patients, this is very common in EMRs.
Meanwhile, the previous works, which ignore the discharge states, only consider admitted ones. In other word, they rarely take the length of stay in each visit into account for patient journeys.

The other limitation is that existing methods rely on a large volume of training data, which is generally not easily available due to both labour-intensive costs and privacy concerns \cite{long2021federated}.
Fortunately, some recent works in natural language processing (NLP) field provide effective solutions when task-specific supervised data is scarce. 
One promising research direction is to leverage off-the-shelter relational knowledge~\cite{zhang2019ernie,liu2020k} (e.g., Freebase and DBpedia) to enhance the model especially when the knowledge can be used as supportive evidence to the targeted task. 
Taking this inspiration, recent works~\cite{Choi_Bahadori_2017_gram,Ma2018-gu_kame,song2019medical} train medical code embeddings upon medical ontology by using graph-based attention mechanism, and thus deliver competitive performance even with insufficient task-specific supervised data. 
Despite their success in several healthcare tasks, these methods still suffer from a main limitation: rich dependency information underlying the patient's sequential EHR is rarely exploited during medical ontology learning. For example, the medical codes from visit information and medical ontology are heterogeneous, and how to effectively learn both representations and fuse their heterogeneous features is a challenging open task. 

To cope with aforementioned limitations, we propose an end-to-end robust transformer-based healthcare analytics model, named \textbf{SE}quential Diagnosis Prediction with \textbf{T}ransformer and \textbf{O}ntological \textbf{R}epresentation (SETOR).
SETOR integrates medical ontology to alleviate data insufficiency, and exploits neural ordinary differential equation (ODE)~\cite{chen2018neural,rubanova2019latent} to tackle temporal irregularity occurred in both two consecutive visits and length of stay in each visit. 
Specifically, SETOR first employs the attention-based graph-embedding approach to learn ontological and generalized representations of medical codes to mitigate the problem of data insufficiency. 
Next, the ontological encoder is proposed to integrate the learned ontological representations into visit information to enhance medical representations. 
Then, the proposed model utilizes neural ODE to learn the discharge state based on the admitted state for each visit, called LoS (Length of Stay) ODE,  and the hidden states for irregular intervals between consecutive patient's visits, called Interval ODE. 
Lastly, SETOR integrates the learned hidden discharge and interval states and compressed visit vectors to predict sequential diagnoses followed by patient journey transformer.
Consequently, the proposed model can improve the prediction quality of future diagnoses, and advance the robustness irrespective of sufficient or insufficient data. 

To summarize, our main contributions are: 
\begin{itemize}
	\item novel LoS and interval ODE representations, that use neural ODE to model discharge states and capture the irregular admitted interval dependencies between patient's visits;
	\item an end-to-end neural network called ``SETOR'' that accurately predicts sequential diagnoses using neural ODE and ontological representation;
	\item an evaluation on two real-world datasets,  qualitatively demonstrating the interpretability of the learned representations of medical codes and quantitatively validating the effectiveness of the proposed model.
\end{itemize}

The remainder of this paper is organized as follows. Section~\ref{sec:rel} reviews related works. 
Then, details about our model are presented in Section~\ref{sec:Method}. And next, in Section~\ref{sec:Experim}, we demonstrate the experimental results conducted on real-world datasets. Lastly, we conclude our work in Section~\ref{sec:Con}.

\section{Related Work}\label{sec:rel}
Deep neural networks have been applied to healthcare analytical tasks, which have recently attracted enormous interest in healthcare informatics. This section reviews two types of related studies, which are sequential prediction and medical ontologies on EHR. 

\subsection{Sequential prediction on EHR}
Sequential prediction of clinical events based on EHR data has been attracting tremendous attentions. 
Most existing models utilize RNNs and attention mechanism for predicting the future diagnoses. 
Med2Vec~\cite{Choi_2016_med2vec} and MIME~\cite{choi2018mime} indirectly exploit an RNN to embed the visit sequences into a patient representation by multi-level representation learning to integrate visits and medical codes based on visit sequences and the co-occurrence of medical codes to predict future health outcomes. 
Other research works have, however, used RNNs directly to model time-ordered patient visits for predicting diagnoses~\cite{choi2016retain,Choi_Bahadori_2017_gram,Ma2017-gs_Dipole,qiao2018pairwise,Ma2018-gu_kame,Ma2018-ao,baytas2017patient,peng2019attentive,Peng2019TemporalSN}.
For example, Dipole~\cite{Ma2017-gs_Dipole} and RETAIN~\cite{choi2016retain} employ RNNs to model the sequential relationships among the medical concepts, guided by future diagnoses prediction task in an end-to-end learning manner. 
Attention-based models, such as, MMORE~\cite{song2019medical} and MusaNet~\cite{peng2020self}, have been employed to capture both visits' dependencies and sequential information in the EHR data to predict future diagnoses. 
Most of these methods rarely make use of the discharge states and irregular intervals in the EHR data. 
Recently, neural ODE~\cite{chen2018neural,rubanova2019latent} has been proposed to handle arbitrary time gaps between observations, which provides an opportunity to alleviate the limitations of the existing models.

\subsection{Medical Ontologies on EHR}
Though healthcare information systems have accumulated huge volumes of EHR data, the data is generally not easily available due to both labour-intensive costs to label training data and privacy concerns of patient data. 
Facing the challenge of insufficient data, additional medical ontologies have been utilized to improve the quality of the medical code embeddings and the predictive performance. 
For instance, GRAM~\cite{Choi_Bahadori_2017_gram} proposes the graph-based attention model to incorporate the medical ontology with an attention mechanism and recurrent neural networks for representation learning with the application to diagnosis prediction.
KAME~\cite{Ma2018-gu_kame} extends GRAM model to additionally consider side information of the learned embedding of the non-leaf nodes in medical ontology, and exploits RNN to integrate the knowledge from both the medical codes and non-leaf nodes and the EHR data to predict future diagnoses. 
MMORE~\cite{song2019medical} learns multiple ontological representations for the non-leaf nodes in the ontology and integrates the EHR co-occurrence statistics to predict sequential diagnoses. 
However, these models do not mutually integrate medical codes and the ontology, leaving learning effective representations from both EHR data and the ontologies an open question.

\begin{figure*}[th]
  \includegraphics[width=1\textwidth]{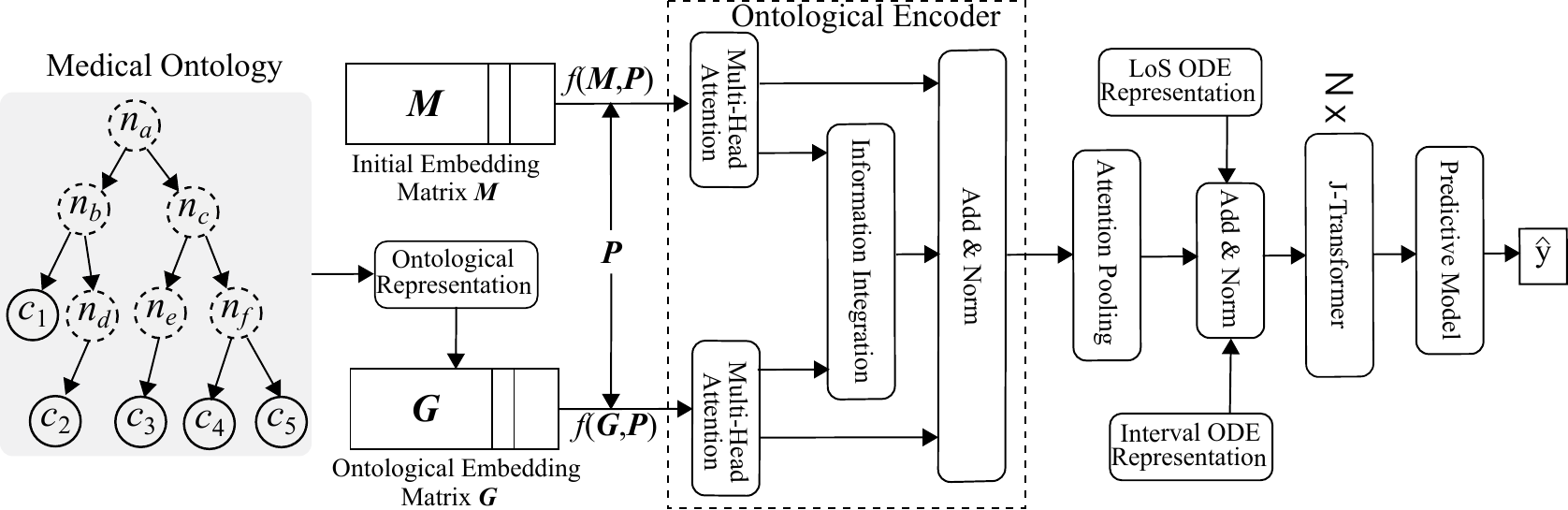}
  \caption{The Proposed SETOR Model. The medical ontology is formatted as a directed acyclic graph, in which, the root node is virtual to construct the tree, the leaf nodes (solid circles) denote fine-grained diagnoses, and the non-leaf nodes (dotted circles) denote coarse-grained disease concepts.}
  \label{fig:model}
\end{figure*}

\section{Methodology}\label{sec:Method}
This section starts with notations of several important concepts and problem statement in the paper. The remainder mainly focuses on details of the proposed model consisting of patient journey transformer, ontological and ODE representations, and task of sequential diagnoses prediction.

\subsection{Notations and Problem Statement}\label{sec:notes}
\subsubsection{Notations}
We denote the set of medical codes from the EHR data as $c_1, c_2, \dots, c_{|\mathbb{C}|} \in \displaystyle \sC$ and $|\displaystyle \sC|$ is the number of unique medical codes. 
Patients' clinical records can be represented by a sequence of visits $\displaystyle \mP=\langle V_1, \dots, V_t, \dots, V_T\rangle$,  which is referred to as the patient journey in the paper, where $T$ is the number of visits in the patient journey.
Each visit $V_t$ consists of a subset of medical codes ($V_t \subseteq \displaystyle \sC$). For clear demonstration, all algorithms will be presented with a single patient's journey.
On the other hand, a medical ontology $\displaystyle \gG$ contains the hierarchy of various medical concepts with the \textit{parent-child} semantic relationship. In particular, the medical ontology $\displaystyle \gG$ is a directed acyclic graph (DAG) and the nodes of $\displaystyle \gG$ consist of leaves and their ancestors, shown in left part in Fig.~\ref{fig:model}. 
Each leaf node refers to a medical code in $\displaystyle \sC$, which is associated with a sequence of ancestors from the leaf to the root of $\displaystyle \gG$. 
And each ancestor node belongs to the set $\displaystyle \sN = {n_{|\mathbb{C}|+1}, n_{|\mathbb{C}|+2}, \dots ,n_{|\mathbb{C}|+|\mathbb{N}|}}$, where $|\displaystyle \sN|$ is the number of ancestor codes in $\displaystyle \gG$. 
A ancestor node in the medical ontology $\displaystyle \gG$ represents a related but more general concept over its children. 
Thus, including these semantic relationships would help the model to improve the medical concept representation that can lead to more accurate predictions of sequential diagnoses. Table~\ref{tab_notes} summarizes notations we will use throughout the paper.

\renewcommand{\arraystretch}{1.1}
\begin{table}[htbp]\small
\caption{Notations for SETOR.}\label{tab_notes}
\centering
	\scalebox{1}{
        \begin{tabular}{|c|l|}
        \hline
        \textbf{Notation} & \textbf{Description} \\
        \hline
        $\displaystyle \sC$      & Set of unique medical codes in dataset \\ \hline
        $|\displaystyle \sC|$     & The number of unique medical codes \\ \hline
        $c_i$     & $c_i \in \displaystyle \sC$, the $i$-th medical code in $\mathbb{C}, \displaystyle i=1,\dots, |\displaystyle \sC|$ \\ \hline
        $V_{t}$ & The \textit{t}-th visit of the patient, $V_t \subseteq \displaystyle \sC$\\ \hline
         $\displaystyle \mP$ & The patient journey, $\displaystyle \mP=\langle V_1, \dots, V_t, \dots, V_T\rangle$  \\ \hline
        $\displaystyle \gG$        & The medical ontology, a directed acyclic graph\\ \hline 
        $\displaystyle \sN$   & Set of ancestor codes in $\displaystyle \gG$\\ \hline
        $\displaystyle \mG$     & Ontological embedding matrix  \\ \hline
       $\displaystyle \mM$ &  Embedding matrix of medical codes\\ \hline
        $\displaystyle \mE_{i,:}$ & Basic embedding vector of medical code $c_i$  \\ \hline
        $d$       & The dimension of medical code embedding\\ 
        \hline
\end{tabular}}
\end{table}

\subsubsection{Problem Statement}
Given a time-ordered patient journey $\displaystyle \mP=\langle V_1, \dots, V_t, \dots, V_T\rangle$, and medical ontology $\displaystyle \gG$, the goal of a sequential diagnosis prediction problem is to predict the next visit information. For the $t$-th visit, where $\displaystyle t=1,2,\dots, T-1$, the outputs are $V_2, V_3, \dots, V_{T}$.

\subsection{Model Overview}
To make the best use of the irregular and temporal properties in EHR and alleviate the challenge of insufficient data, we propose a robust and transformer-based model, called SETOR, illustrated in Fig.~\ref{fig:model}. First, the medical ontology $\displaystyle \gG$ is embedded into an ontological embedding matrix $\displaystyle \mG$. 
Then, an ontological encoder aggregates both the embedded diagnoses from $\bm{G}$ and an initial embedding matrix $\bm{M}$ by embedding operation $\displaystyle f(\cdot,\displaystyle \mP)$ to learn both co-occurrence and medical knowledge. 
$\displaystyle \mM \in\mathbb{R}^{|\displaystyle \sC| \times d}$ is an embedding matrix of medical codes, where $d$ represents the embedding size. $\displaystyle \mM$ is randomly initialised by a uniform distribution, and its entries are learnable during model training in an end-to-end manner.
The outputs of the ontological encoder are fed into attention pooling layer to compress a set of medical codes in a visit into a vector representation. 
Next, our proposed LoS and Interval ODE representations are added to the learned visit representations, and the normalized outputs are fed into a journey transformer to learn the visit dependencies in a patient journey. Lastly, a predictive model, following the journey transformer, is used to predict the next visit information.

\subsection{Ontological Representation}\label{sec:o-rep}
To mitigate the problem of data insufficiency in healthcare and to learn knowledgeable and generalized representations of medical codes, we employ the attention-based graph representation approach GRAM~\cite{Choi_Bahadori_2017_gram}. 
In the medical ontology $\displaystyle \gG$, each leaf node $c_i$ has a basic learnable embedding vector $\displaystyle \mE_{i,:} \in \mathbb{R}^d$, where $1 \leq i \leq |\mathcal{C}|$, and \textit{d} represents the dimensionality. 
And each non-leaf node $n_i$ also has an embedding vector $\displaystyle \mE_{i,:} \in \mathbb{R}^d$, where $|\displaystyle \sC|+1 \leq i \leq |\displaystyle \sC|+|\displaystyle \sN|$. $\displaystyle \mE$ is initialized with the values from the uniform distribution, and $\displaystyle \mE \in \mathbb{R}^{(|\displaystyle \sC|+|\displaystyle \sN|) \times d}$.
The attention-based graph embedding uses an attention mechanism to learn the \textit{d}-dimensional final embedding $\displaystyle \mG$ for each leaf node \textit{i} (medical code) via:
\begin{equation}\label{eq:g-embedding}
  \displaystyle \mG_{i,:}=\sum_{j\in \displaystyle \parents_\gG(i)}\alpha_{ij}\displaystyle \mE_{j,:}
\end{equation}
where $\displaystyle \parents_\gG(i)$ denotes the set comprised of leaf node \textit{i} and all its ancestors, $\displaystyle \mE_{j,:}$ is the \textit{d}-dimensional basic embedding of the node \textit{j} and $\alpha_{ij}$ is the attention weight on the embedding $\displaystyle \mE_{j,:}$ when calculating $\displaystyle \mG_{i,:}$, which is formulated by following the Softmax function,
\begin{equation}
\alpha_{ij} = \frac{\exp(g(\displaystyle \mE_{i,:}, \displaystyle \mE_{j,:}))}{\sum_{k\in \displaystyle \parents_\gG(i)}\exp(g(\displaystyle \mE_{i,:}, \displaystyle \mE_{k,:}))}.
\end{equation}
\begin{equation}
g(\displaystyle \mE_{i,:}, \displaystyle \mE_{j,:}) = \bm{w}_{\alpha}^T \texttt{tanh} \left(\displaystyle \mW_{\alpha}(\displaystyle \mE_{i,:} || \displaystyle \mE_{j,:}) + \displaystyle \vb_{\alpha}\right),
\end{equation}
where $(\displaystyle \mE_{i,:}|| \displaystyle \mE_{j,:})$ is to concatenate $\displaystyle \mE_{i,:}$ and $\displaystyle \mE_{j,:}$ in the child-ancestor order; $\bm{w}_{\alpha}$, $\displaystyle \mW_{\alpha}$ and $\displaystyle \vb_{\alpha}$ are learnable parameters.

\subsection{Ontological Encoder}\label{sec:o-encoder}
To encode both visit information and medical ontology as well as fuse their heterogeneous features, we propose ontological encoder. The dotted rectangle in Fig.~\ref{fig:model} shows the details of the encoder, where. We first calculate the code embeddings $\displaystyle \tE^M=f(\bm{M}, \bm{P})$
and node embedding $\displaystyle \tE^G=f(\bm{G}, \bm{P})$, where $\displaystyle \tE^M, \displaystyle \tE^G \in \mathbb{R}^{(T-1) \times n \times d}$ are 3-dimensional tensors, the function $f$ is to embed medical codes in patient journey $\bm{P}$ according to $\bm{M}$ and $\bm{G}$. Next, $\displaystyle \tE^M, \displaystyle \tE^G$ are fed into two different multi-head self-attentions ($\texttt{MultiHead}$)~\cite{vaswani2017attention}, where $n$ is the number of medical codes in each visit of the patient journey. For simplicity, we take the $t$-th visit ($\displaystyle t=1,2,\dots, T-1$) in patient journey $\bm{P}$ as an example to demonstrate the process of ontological encoder as follow, 
\begin{equation}
\label{e_code2visit}
\begin{aligned}
\bm{V}_{Mt} = \texttt{MultiHead}(\displaystyle \tE^m_{t,:,:},\displaystyle \tE^m_{t,:,:},\displaystyle \tE^m_{t,:,:}),\\
\bm{V}_{Gt} = \texttt{MultiHead}(\displaystyle \tE^G_{t,:,:},\displaystyle \tE^G_{t,:,:},\displaystyle \tE^G_{t,:,:}).
\end{aligned}
\end{equation}
where $\texttt{MultiHead}$ is a function of multi-head attention~\cite{vaswani2017attention}, and $\bm{V}_{Mt}, \bm{V}_{Gt} \in \mathbb{R}^{n \times d}$.
\subsubsection{Multi-Head Attention}\label{sec:multi-head}
The multi-head attention mechanism relies on self-attention, where all of the keys, values and queries come from the same place. The self-attention operates on a query $\displaystyle \mQ$, a key $\displaystyle \mK$ and a value $\displaystyle \mV$:
\begin{equation}
\mathrm{Attention}(\displaystyle \mQ,\displaystyle \mK,\displaystyle \mV) = \mathrm{softmax}(\frac{\displaystyle \mQ \displaystyle \mK^T}{\sqrt{d}})\displaystyle \mV
\end{equation}
where $\displaystyle \mQ$, $\displaystyle \mK$, and $\displaystyle \mV$ are $n \times d$ matrices, $n$ denotes the number of medical codes in a visit in a patient journey, $d$ denotes the dimension of embedding.

The multi-head attention mechanism obtains $h$ (i.e. one per head) different representations of ($\displaystyle \mQ,\displaystyle \mK,\displaystyle \mV$), computes self-attention for each representation, concatenates the results. This can be expressed as follow:
\begin{equation}
\label{head}
\mathrm{head}_i = \mathrm{Attention}(\displaystyle \mQ \displaystyle \mW^Q_i, \displaystyle \mK \displaystyle \mW^K_i, \displaystyle \mV \displaystyle \mW^V_i)
\end{equation}
\begin{equation}
\label{multi-head}
\mathrm{MultiHead}(\displaystyle \mQ,\displaystyle \mK,\displaystyle \mV) = \mathrm{Concat}(\mathrm{head}_1,...,{head}_h)\displaystyle \mW^O
\end{equation}
where the projections are parameter matrices $\displaystyle \mW^Q_i \in \mathbb{R}^{d\times d_k}, \displaystyle \mW^K_i \in \mathbb{R}^{d\times d_k}, \displaystyle \mW^V_i \in \mathbb{R}^{d\times d_v}$ and $\displaystyle \mW^O \in \mathbb{R}^{hd_v\times d}$, $d_k = d_v = d/h$.

\subsubsection{Information Integration}
The ontological encoder adopts an information integration layer for the mutual integration of the code and node embedding in a visit. The process is as follows:
\begin{equation}
\begin{aligned}
	\bm{H}_{t} &= \sigma (\bm{W}_{M} \bm{V}_{Mt} + \bm{W}_{G} \bm{V}_{Gt} + \bm{b})
\end{aligned}
\end{equation}
where $\bm{W}_{M}, \bm{W}_{G}, \bm{b}$ are learnable parameters, $\bm{H}_t \in \mathbb{R}^{n \times d}$ is the inner hidden state integrating the information of both the code and the node. $\sigma(\cdot)$ is the non-linear activation function, which usually is the ReLU function.

The output of ontological encoder is denoted as follows, 
\begin{equation}
\begin{aligned}
\bm{O}_t = \texttt{LayerNorm} ( \bm{H}_{t} + \bm{V}_{Mt} + \bm{V}_{Gt}),
\end{aligned}
\end{equation}
where  $\bm{O}_t \in \mathbb{R}^{n \times d}, ~1 \leq t \leq (T-1)$, so that we can represent heterogeneous information of medical codes and ontology into a united feature space.

For ontological representation and encoder, we first build an embedding matrix for both leaf and non-leaf nodes in medical ontology. 
Then, we extract knowledge for each code from medical ontology as a tuple $(leaf,~ancestors)$, 
and embed each code in the tuple by looking up the embedding matrix. 
Lastly, we calculate knowledge-enriched representation of each leaf node using Eq.(1). 
During this procedure, we also consider interactively encoding both visiting records and medical ontology by presenting an ontological encoder to fuse their heterogeneous features.

\subsection{Attention Pooling}\label{sec:attn_pool}
Attention Pooling~\cite{lin2017structured,cai2018medical} explores the importance of each individual code within a visit. It works by compressing a set of medical code embeddings from the visit into a single context-aware vector representation. For simplicity, we take the $t$-th transformer output $\bm{V}^t$ as an example. Formally, it is written as:
\begin{equation}
g(\bm{V}^t_{i,:}) = \displaystyle \vw^T \sigma (\displaystyle \mW^{(1)}\bm{V}^t_{i,:} + b^{(1)})+b,
\end{equation}
where $\bm{V}^t_{i,:}$ is the $i$-th row of $\bm{V}^{t}$ ($1 \leq i \leq n$ ), $\sigma$ is ReLU function and $\displaystyle \vw, \displaystyle \mW^{(1)}, \displaystyle \vb^{(1)}, \displaystyle \vb$ are learnable parameters. The probability distribution is formalized as
\begin{equation}
\label{alpa-add}
\bm{\alpha}_t = \texttt{softmax}([g(\bm{V}^t_{i,:})]_{i=1}^n).
\end{equation}

The final output $\bm{v}_t$ of the attention pooling is the weighted average of sampling a code according to its importance, i.e., 
\begin{equation}
\label{eq:v-embedding}
\bm{v}_t = \sum_{i=1}^n \bm{\alpha}_t \odot [\bm{V}^t_{i,:}]_{i=1}^n,
\end{equation}
where $\bm{v}_t \in \mathbb{R}^{d} ~(1 \leq t \leq (T-1))$ represents the $t$-th visit in the patient journey.

\subsection{ODE Representations}\label{neural-ode-section}
Neural ODE~\cite{chen2018neural,rubanova2019latent,dupont2019augmented} models the time series as a continuously changing trajectory and makes better use of the data’s timestamp information and predictions arbitrarily in time. 
Each trajectory is determined by the local initial state $\displaystyle \vh_{t_{1}}$ and the global set of potential dynamics shared by the all-time series. Given observation time $t_{1}$, $t_{2}$,..., $t_{T}$ and initial state $\displaystyle \vh_{t_{1}}$, generated by the ODE solver $\displaystyle \vh_{t_{1}}$ ,..., $\displaystyle \vh_{t_{(T-1)}}$, which describes the underlying state of each observation. 
This generation model can be defined by the following formula:
\begin{gather} 
\frac{\partial \displaystyle \vh(t)}{\partial t} = f(\displaystyle \vh(t), \bm{\theta}_f),~ \text{where}~\displaystyle \vh(t_1) = \displaystyle \vh_{t_{1}}, \\ 
\displaystyle \vh_{t_{1}},...,\displaystyle \vh_{t_{T-1}} = \texttt{ODESolve }(\displaystyle \vh_{t_{1}},f,\bm{\theta}_{f},t_{1},...,t_{T-1}),
\label{eq:ode}
\end{gather}
where 
$f$ is a time-invariant function, using a neural network with parameters $\bm{\theta_f}$. 
The function takes the value $\displaystyle \vh$ at the present time step and outputs the gradient at the end. 

ODE is a function as $h(t)$ in Eq.(11). 
The equation needs to be solved during each evaluation and begins with an initial state $h_{t_1}$, which is also called initial value problem. 
Adjoint method by Pontryagin is employed to calculate the gradients of the ODE. With a low memory footprint, this method works by solving second, augmented ODE backwards in time and can be used with all ODE integrators.
By solving the equation, the desirable sequence of hidden states can then be produced for downstream modules. 

\textbf{LoS ODE Representation.} As each visit has two timestamps (admitted and discharge times) and each initial state (e.g. the $i$-th $\bm{v}_i, i=1,.., (T-1)$) is given by the output of attention pooling, we can utilize Neural ODE to predict the discharge state as follow:
\begin{equation}
    \bm{v}_i,\bm{v}^{dis}_i = \texttt{ODESolve }(\bm{v}_i,f,\bm{\theta}_{f},t_i,t^{dis}_i),
\end{equation} 
where $\bm{v}^{dis}_i$ is the discharge state for the $i$-th visit, and $t_i,t_i^{dis}$ are the admitted and discharge timestamps,respectively.

\textbf{Interval ODE Representation.} A patient journey consists of a sequence of irregular visits with timestamps. We can utilize Neural ODE to learn the hidden state for each visit timestamp following Equation~\ref{eq:ode}.
\begin{gather} 
\displaystyle \vh_{{1}} \sim p(\bm{v}_{1}), \\ 
\displaystyle \vh_{{1}},...,\displaystyle \vh_{{T-1}} = \texttt{ODESolve }(\displaystyle \vh_{{1}},f,\bm{\theta}_{f},t_{1},...,t_{T-1}),
\end{gather}
where $p$ is a probability distribution dependent on time, and $\bm{v}_1$ is the first admitted visit state of a patient journey. The outputs of LoS and Interval ODE representations are added to the outputs of attention pooling and normalized for the following layer, which is denoted as follows: 
\begin{equation}
\small
\begin{aligned}
\tilde{\bm{v}}_t = \texttt{LayerNorm} ( \bm{v}_{t} + \bm{v}^{dis}_t + \displaystyle \vh_{{t}}),~1 \leq t \leq (T-1).
\end{aligned}
\end{equation}

\subsection{Patient Journey Transformer Module}\label{sec:res-trans}
To learn visits' relationships in a patient journey, a module, called $\texttt{J-Transformer}$, is proposed to capture inherent dependencies, as shown in Fig.~\ref{fig:model}.
$\texttt{J-Transformer}$ responsible for learning dependencies of sequential visits in a patient journey with admission intervals and length of stay in each visit, which is calculated as follows:
\begin{equation}
\small
\begin{aligned}
\{\bm{v}^o_{1}, \ldots, \bm{v}^o_{T-1}\} = \texttt{J-Transformer} (\{\bm{o}_{1}, \ldots, \bm{o}_{T-1}\}).
\end{aligned}
\end{equation}

\begin{figure}[th]
\centering
  \includegraphics[width=0.35\textwidth]{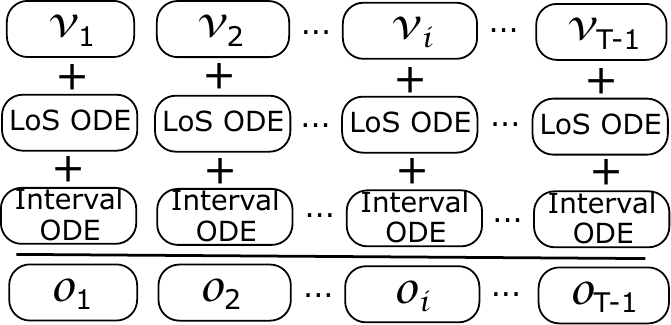}
  \caption{Detailed inputs to $\texttt{J-Transformer}$}
  \label{fig:trans_input}
\end{figure}

Besides, we denote the number of $\texttt{J-Transformer}$ layers as $N$. $\bm{o}_{t}$ is the input to $\texttt{J-Transformer}$, which is depicted in Fig.~\ref{fig:trans_input}. $\texttt{J-Transformer}$ is identical to its implementation in BERT~\cite{vaswani2017attention} and~\cite{Devlin2018-ae}, which has two sub-layers. The first is a multi-head attention mechanism mentioned in Section~\ref{sec:multi-head}, and the second is position wise fully connected feed-forward network. Residual connection~\cite{he2016deep} is employed around each of the two sub-layers, followed by layer normalization~\cite{ba2016layer}.

\subsection{Sequential Diagnoses Prediction}
Given a patient's visit records $\bm{P} = \{V_1, V_2, \ldots, V_{T-1}\}$, to capture the EHR sequential visit information we perform the sequential diagnoses predictive task with the objective of predicting the disease codes of the next visit $V_{t}$,  which can be expressed as follows:
\begin{gather}
\label{eq:seq_loss}
\begin{aligned}
 \hat{\bm{y}}_{t} = \texttt{Softmax}(\bm{W}\bm{v}^o_{t-1} + \bm{b}),&
\end{aligned}\\
\small
\begin{aligned}
\mathcal{L} = \frac{1}{T}\sum_{t=2}^{T-1}-\left({\bm{y}_t}^{\texttt{T}}\log{\hat{\bm{y}}}_t+(1-{\bm{y}_t})^{\texttt{T}}\log{(1-\hat{\bm{y}}_t)}\right),
\end{aligned}
\end{gather}
where $\bm{v}^o_{t-1}\in \mathbb{R}^d$ is the output of $\texttt{J-Transformer}$ to denote the representation of the ($t-1$)-th visit, $\mathcal{L}$ is the loss function, $\bm{y}_t$ is a vector with $|\displaystyle \sC|$ elements, whose value is 1 if the $i$-th diagnosis code exists in $V_t$ and 0 otherwise, and $\bm{W} \in \mathbb{R}^{|\displaystyle \sC|\times d}$ and $\bm{b} \in \mathbb{R}^{|\displaystyle \sC|}$ are the learnable parameters.

\section{Experiments}\label{sec:Experim}

In this section, we conduct experiments on two real world medical claim datasets to evaluate the performance of the proposed SETOR. Compared with the state-of-the-art predictive models, SETOR yields better performance on different evaluation strategies.

\subsection{ Data Description}

We conducted comparative studies on two real-world datasets in the experiments, which are the MIMIC-III~\cite{Johnson_2016} and MIMIC-IV~\cite{MIMIC-IV} databases. 

\paragraph{M-III Dataset}
The MIMIC-III dataset~\cite{Johnson_2016} is an open-source, de-identified dataset of ICU patients and their EHRs between 2001 and 2012. The diagnosis codes in the dataset follow the ICD9 standard. MIMIC-III is denoted by M-III in the experiment.

\paragraph{M-IV Dataset}
The MIMIC-IV dataset~\cite{MIMIC-IV} is an update to MIMIC-III, which incorporates contemporary data and improves on numerous aspects of MIMIC-III. The dataset consists of the medical records of 73,452 patients between 2008 and 2019. MIMIC-IV is denoted by M-IV.

Tab.~\ref{tab:stats} shows the statistical details about the datasets, where the selected patients made at least two visits. MIMIC-III and MIMIC-IV are denoted by M-III and M-IV in the experiment, respectively.

\renewcommand{\arraystretch}{1.2}
\begin{table}
  \centering
  \caption{Statistics of the datasets.}
  \scalebox{1}{%
      \begin{tabular}{|l|r|r|}
        \hline
        \textbf{Dataset}&\textbf{M-III}&\textbf{M-IV}\\
        \hline
        \# of patients  & 7,499&73,452 \\
        \# of visits  &  19,911&295,351\\
        Avg. \# of visits per patient  & 2.66&4.02\\
        \hline
        \# of unique ICD9 codes  &  4,880&9,165\\
        Avg. \# of ICD9 codes per visit & 13.06&12.01\\
        Max \# of ICD9 codes per visit& 39&57\\
        \hline
        \# of category codes&272&283 \\
        Avg. \# of cat. codes per visit&11.23&10.41\\
        Max \# of cat. codes per visit&34&37\\
      \hline
    \end{tabular}
    }
  \label{tab:stats}
\end{table}

\subsection{Experimental Setup}
In this subsection, we first introduce the state-of-the-art approaches for diagnosis prediction task in healthcare, and then outline the measures used for predictive performance evaluation. Finally, we describe the implementation details.

\paragraph{Baseline Approaches}
We compare the performance of our proposed model against the following state-of-the-art baseline models: 
\begin{itemize}
\item \textbf{RETAIN}~\cite{choi2016retain}, which learns the medical concept embeddings and performs heart failure prediction via the reversed RNN with the attention mechanism.
\item \textbf{Dipole}~\cite{Ma2017-gs_Dipole}, which uses bidirectional RNN and three attention mechanisms (location-based, general, concatenation-based) to predict patient visit information. We chose location-based Dipole as a baseline method.
\item \textbf{GRAM}~\cite{Choi_Bahadori_2017_gram}, which is a graph-based attention model to learn the representations from the knowledge graph to predict future medical outcomes. .
\item \textbf{KAME}~\cite{Ma2018-gu_kame}, which is a diagnosis prediction model inspired by GRAM, using medical ontology to learn representations of medical codes and their parent codes. These are then used to learn input representations of patient data which are fed into a Neural Network architecture to predict sequential diagnoses.
\item \textbf{MMORE}~\cite{song2019medical}, which is based on medical ontology with an attention mechanism.
\end{itemize}

\paragraph{Predictive Task}

The purpose of the sequential diagnosis prediction task is to predict the diagnosis information of the next visit. In the experiments, true labels $\bm{y}_t$ are prepared by grouping the ICD9 codes into 283 groups using CCS single-level diagnosis grouper\footnote{https://www.hcup-us.ahrq.gov/toolssoftware/ccs/AppendixASingleDX.txt}. It is to improve the training speed and predictive performance, while preserving sufficient granularity for all the diagnoses. 
We measure the predictive performance by $Accuracy@k$, which are defined as:
\begin{align*}
\small
    Accuracy@k = \frac{\text{\# of true positives in the top \textit{k} predictions}}{\text{\# of positives}}
\end{align*}

Sequential diagnosis prediction is a multi-label problem, so normal Accuracy is inapplicable. 
Following previous works, we also use accuracy@$k$ as the metric, which measures the ratio of positive labels ranked in top-$k$ according to their logits. Specifically given a test sample, we first calculate the logits for all categories by a trained model, and rank the categories by the logits in descending order. Then, we count how many positive labels fall into top-$k$ and compute the ratio over the number of all positive labels in this sample.
Lastly, we derive the final metric, Accuracy@$k$, by averaging the ratios on the samples from the entire test set.

\paragraph{Implementation Details}
We use CCS-multi-level diagnoses hierarchy as the medical ontology. We implement all the approaches with Pytorch 1.4.0. For the training models, we use Adadelta~\cite{zeiler2012adadelta} with a minibatch of 32 patients. We randomly split the data into a training set, validation set and test set and fix the size of the validation set to be 10\%. 
To validate the robustness against insufficient data, we vary the size of the training set from 20\% to 80\% and use the remaining part as the test set. 
The validation set is used to determine the best parameters values in the 100 training iterations. The drop-out strategies (the drop-out rate is 0.1) are used for all the approaches. We set dimension $d = 200$ for all the baselines and the proposed model. 


\begin{figure*}[htbp]
    \centering
     \includegraphics[width=0.8\textwidth]{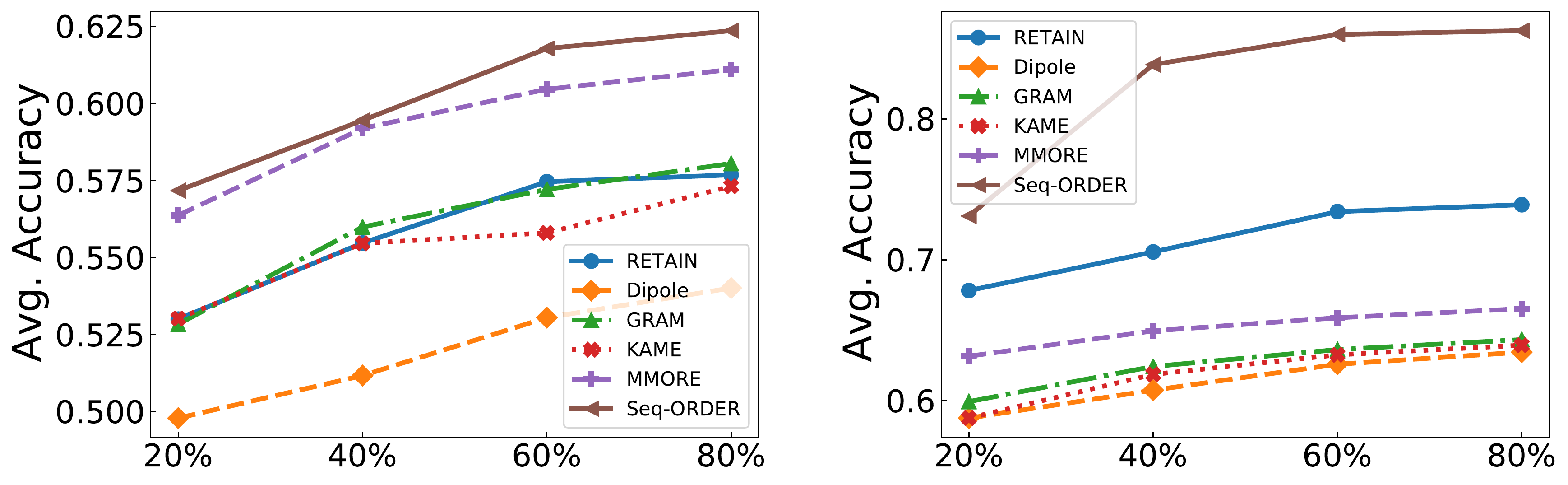}
     \begin{subfigure}[b]{0.54\textwidth}
         \centering
         \caption{M-III}\label{sub-fig:mimic}
     \end{subfigure}
     \hfill
     \begin{subfigure}[b]{0.45\textwidth}
         \centering
         \caption{M-IV}\label{sub-fig:m-iv}
     \end{subfigure}
  \caption{Accuracy@20 of diagnoses prediction on M-III and M-IV, size of training data is varied from 20\% to 80\%.}
  \label{fig:mimic-eicu}
\end{figure*}

\subsection{Results of Sequential Diagnosis Prediction}
Tab.~\ref{tab:Accuracy} shows the accuracy@\textit{k} of the SETOR and baselines with different \textit{k} on two real-world datasets for the sequential diagnoses prediction task. From Tab.~\ref{tab:Accuracy}, we
can observe that the performance of the proposed SETOR is better than
that of all the baselines on the two datasets.

On the M-III dataset, compared with the best baseline MMORE, the accuracy of SETOR improves by 2.21\%. These results suggest that adding LoS and Interval ODE representation layers when predicting diagnoses is effective. We observe that the performance obtained by the models using ontologies is better than that obtained by the models without using ontologies on M-III, which can be thought of as a small and insufficient dataset. The underlying reason is that the models integrating external medical ontologies can alleviate the issue of insufficient data.

On the M-IV dataset, the proposed SETOR still outperforms all the state-of-the-art diagnosis prediction approaches. 
Compared with the best baseline RETAIN, the accuracy of SETOR improves by  8.77\% when $k = 5$. We also find that when not using ontologies RETAIN outperforms the other baseline models on M-IV the large dataset. 
This implies that the model can obtain comparative performance without using ontologies when the size of training dataset is larger. Compared to the models based on GRU and attention (e.g., RETAIN and Dipole), although Dipole fuses location attention, its performance is inferior to the attention-based models.
Overall, our proposed framework exhibits better predictive power for both sufficient and insufficient datasets.

On the two datasets, the results show that the proposed model outperforms all the baselines, especially when the size of dataset is large. This demonstrates that the superiority of SETOR results from the explicit consideration of both the ontologies and the EHR co-occurrence, with the irregular intervals and discharge states being well handled. 

\begin{table}[t]
	\centering
	\caption{Accuracy comparison of sequential diagnoses prediction.}
	\scalebox{1}{%
		\begin{tabular}{|l|l|c|c|c|c|}
			\hline
			\textbf{Dataset} & \textbf{Model}&\multicolumn{4}{c|}{\textbf{Accuracy@k (\%)}} \\
			\cline{3-6} &  &\textbf{5} & \textbf{10} & \textbf{20} & \textbf{30} \\ \hline
			&RETAIN    &27.15&41.41&57.68&68.25\\ 
			&Dipole    &24.55&37.04&54.01&60.09\\ 
			M-III&GRAM &27.72&41.24&58.05&68.08\\ 
			&KAME      &27.98&41.81&57.31&68.02\\ 
			&MMORE     &28.97&43.74&61.10&71.61\\ 
			&SETOR&    \textbf{31.18}&\textbf{45.80}&\textbf{62.36}&\textbf{72.46}\\ \hline
			&RETAIN    &38.95&57.60&73.91&81.48\\ 
			&Dipole    &32.48&47.75&63.45&72.39\\ 
			M-IV&GRAM  &33.63&48.84&64.34&73.05\\ 
			&KAME      &33.56&48.80&63.94&72.64\\ 
			&MMORE     &34.21&50.59&66.53&75.21\\ 
			&SETOR&    \textbf{47.72}&\textbf{71.21}&\textbf{86.26}&\textbf{90.32}\\
			\hline
		\end{tabular}
	}
	\label{tab:Accuracy}
\end{table}

\subsection{Data Sufficiency Analysis}

To analyze the influence of data sufficiency on the predictions, we randomly split the data into training, validation, and test sets, and fix the size of the validation set to be 10\%. To validate the robustness against insufficient data, we vary the size of the training set to form four groups: 20\%, 40\%, 60\%, and 80\%, and use the remaining part as the test set. The training set in the 20\% group are the most insufficient for training the proposed and baseline models, while the data in the 80\% group are the most sufficient for training the models. Fig.~\ref{fig:mimic-eicu} shows the Accuracy@20 on the M-III and M-IV. 

From the Fig.~\ref{sub-fig:mimic}, we can observe that the accuracy of the proposed model is higher than that of the baselines in all groups. Specifically, MMORE is a comparative model with ontological representation and diagnosis co-occurrence over M-III, which shows that the models integrating medical ontology learns reasonable medical code embeddings to improve the prediction with insufficient data. The performance of Dipole is inferior to other baseline models, which indicates that the model only taking the sequential information into accounts is not enough.

When training data on the M-IV, which is a sufficient dataset, Fig.~\ref{sub-fig:m-iv} shows that the proposed model significantly outperforms all the baselines.
We observe that the performance obtained by RETAIN without using medical ontologies is better than that of the other baselines over M-IV. 
The underlying reason may be that the next-admission diagnosis prediction is more sensitive to the diagnosis co-occurrence and the sequential positions of the visits in sufficient data. Overall, the results demonstrate that the proposed model balances medical ontology and diagnosis co-occurrence over both insufficient and sufficient EHR data to further improve prediction performance.

 \begin{figure*}[t]
 \centering
  \includegraphics[width=0.85\textwidth]{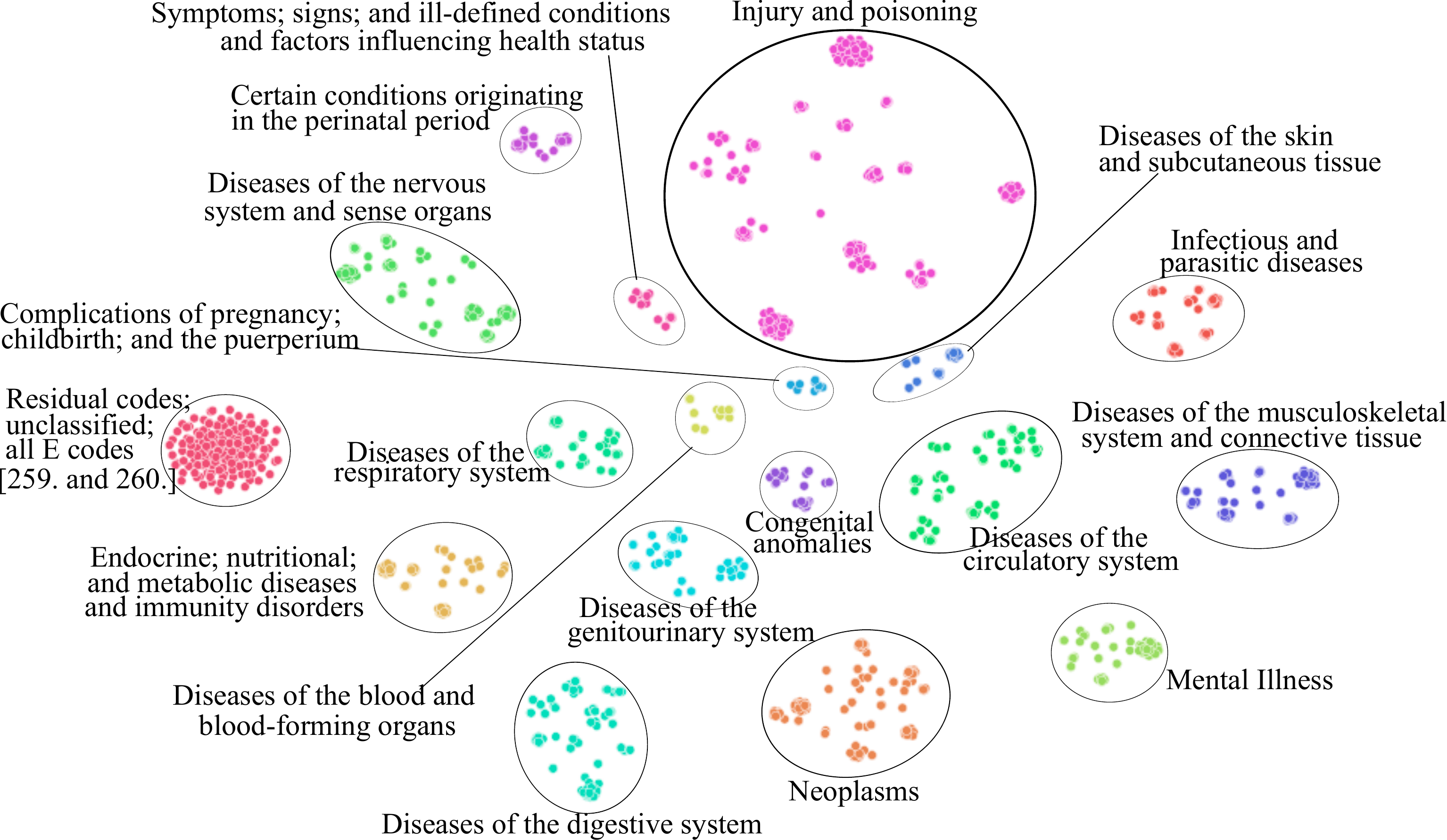}
  \caption{Annotations of SETOR Diagnosis Embedding }
  \label{fig:annot}
\end{figure*}

\begin{figure*}[ht]
  \centering
  \includegraphics[width=0.85\textwidth]{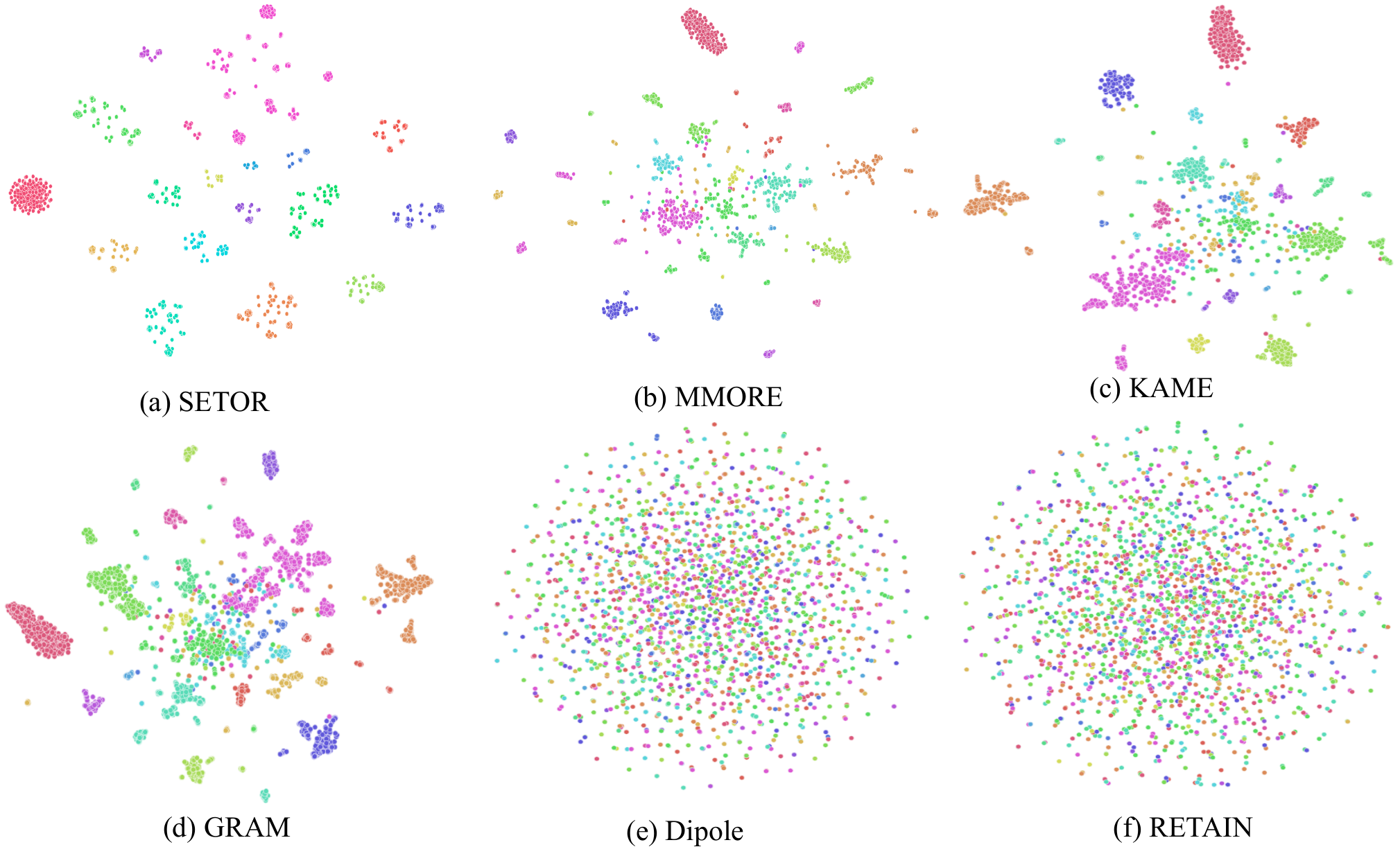}
  \caption{\textit{t}-SNE Scatterplots of Medical Codes Learned by Predictive Model on the M-III dataset.}
  \label{fig:t-sne}
\end{figure*}

\subsection{Ablation Study}
We performed a detailed ablation study to examine the contributions of 
the model's components to the prediction task. There are three components: (Transformer) the transformer blocks to learn the patient journey from the embedded visits; (Ontology) the external medical ontology integrated into SETOR, and (ODE Representations) the LoS and interval encodings to be added to the learned visit embeddings. 
\begin{itemize}
    \item  \textbf{w/o J-Trans:} remove the patient journey transformer blocks from the proposed model; 
	\item  \textbf{w/o Ontology:} remove the ontological representation from the proposed model; 
	\item  \textbf{w/o LoS:} remove LoS ODE representation; 
	\item  \textbf{w/o Interval:} remove the interval ODE representation;
	\item  \textbf{w/o ODE:} replace two ODE representations with position embedding.
\end{itemize}

\paragraph{Ablated Transformers}
$\texttt{J-Transformer}$ is responsible for learning dependencies of sequential visits in a patient journey with admission intervals and length of stay in each visit. We conducted a group of experiments to analyze the contribution of this component to SETOR over two datasets.
From Tab.~\ref{tab:ablation}, we observe that the full complement of SETOR achieved superior accuracy to the ablated models. Specifically, we note that the $\texttt{J-Transformer}$ (w/o J-Trans) contributes the highest accuracy to the predictive task over M-III. Specifically, the accuracy improves by 4.14\% and 4.89\% when $k = 5$ and 20, respectively. On M-IV dataset, the performance of sequential diagnosis prediction is improved further with component of  $\texttt{J-Transformer}$. The accuracy increases by 9.86\% and 5.49\% when $k = 5$ and 20, respectively. The underlying reason is that transformer blocks are providing significant amount of additional parameters and thus capacity. Thus, the prediction performance is significantly improved.

\paragraph{Ablated Representations}
In the paper, representations consists of ontological, LoS and interval representation. From Tab.~\ref{tab:ablation}, we see that the components of various representations have contributions to the proposed model SETOR, though the contributions are no more than that of the $\texttt{J-Transformer}$. Specifically, we observe that the ontological representation (w/o Ontology) contributes the highest accuracy to the predictive task over M-III, which gives us confidence in using external medical ontologies to enhance the patient journey representations without sufficient data. Moreover, it is clear that the component of ODE representations provides valuable information for the performance of sequential diagnoses prediction over M-IV, which implies the irregular intervals and discharge states play more important roles with sufficient data. 
As shown in the ablation study (Tab.~\ref{tab:ablation}), the ontological and ODE representations contribute most to the predictive tasks, no matter the training data is sufficient or not, e.g., 1.18\% lift of Acc@20 on M-IV and 0.56\% lift of Acc@20 on M-III. 
Also as shown in Fig.~\ref{fig:annot}, the embeddings produced by our proposed model shows the great inseparability of disease categories.

\begin{table}[t]
	\centering
	\caption{Ablation Performance Comparison.}
	\scalebox{1}{%
		\begin{tabular}{|l|c|c|c|c|}
			\hline
			\textbf{Ablation}&\multicolumn{2}{c|}{\textbf{M-III} (\%)}&\multicolumn{2}{c|}{\textbf{M-IV} (\%)} \\
			\cline{2-3} \cline{4-5} &\textbf{Acc@5} & \textbf{Acc@20} & \textbf{Acc@5} & \textbf{Acc@20} \\ \hline
			SETOR&    \textbf{31.18}&\textbf{62.36}&\textbf{47.72}&\textbf{86.26}\\
			\hline
			w/o J-Trans     &27.04&57.47&37.86&70.77\\ 
			\hline
			w/o Ontology    &30.39&61.80&47.65&86.07\\  
			w/o LoS         &30.74&61.96&47.59&86.20\\ 
			w/o Interval    &30.95&61.84&47.57&86.18\\ 
			w/o ODE         &30.45&61.93&47.38&85.08\\ 
			\hline
		\end{tabular}
	}
	\label{tab:ablation}
\end{table}

\subsection{Interpretable Representation Analysis}\label{sec:interpret}

To qualitatively demonstrate the interpretability of the learned medical code embeddings by all the predictive models on the M-III dataset, we randomly select 2000 medical codes and then plot on a 2-D space with t-SNE~\cite{maaten2008visualizing} as shown in Fig.~\ref{fig:annot} and Fig.~\ref{fig:t-sne}. Each dot represents a diagnosis code, and the color of the dots represents the 18 disease categories in CCS multi-level hierarchy\footnote{https://www.hcup-us.ahrq.gov/toolssoftware/ccs/AppendixCMultiDX.txt} and the text annotations represent the detailed disease categories. Ideally, the dots with the same color should be in the same cluster and there are margins among different clusters.

From Fig.~\ref{fig:annot}, we can observe that SETOR learns interpretable disease representations that are in accord with the hierarchies of the given medical ontology $\mathcal{G}$, and obtains the 18 non-overlapping clusters.  Specifically, for category of ``Residual codes; unclassified; all E codes [259. and 260.]'', we observe the medical codes in this category are closely clustered together, with large margin to other categories. The embedding results of medical codes in this category are harmony to CCS multi-level hierarchy, as category of ``Residual codes; unclassified; all E codes [259. and 260.]'' has not sub-category in CCS ontology. However, we note that the medical codes in category of ``Injury and poisoning'' are scattered in larger area. The underlying reason is that there are 12 sub-categories under this category. Thus, it is demonstrated that our proposed model SETOR learns meaningful and semantic representations for medical codes, which have practical interpretability.

As shown in Fig.~\ref{fig:t-sne}, compared with SETOR, MMORE, KAME and GRAM are comparative baseline models, as those integrate medical ontology to predict sequential diagnoses. We observe the three baseline models learn reasonably interpretable diagnosis representations for several categories, as there is a large number of dots over-lapping in the center part of Fig.~\ref{fig:t-sne}b, \ref{fig:t-sne}c, and \ref{fig:t-sne}c. It is clear that the medical codes in category of ``Residual codes; unclassified; all E codes [259. and 260.]'' are well clustered. But for the medical codes in category of ``Injury and poisoning'', their learned embeddings do not have clear margins to other categories.    Fig.~\ref{fig:t-sne}e and \ref{fig:t-sne}f suggest that models not using medical ontologies cannot easily learn interpretable representations. In addition, the predictive performance of SETOR is much better than that of MMORE, KAME, and GRAM shown in Tab.~\ref{tab:Accuracy}, which proves that the proposed model does not affect the interpretability of medical codes. Moreover, it effectively improves the prediction accuracy.

\section{Conclusion}\label{sec:Con}
Although the recent approaches have achieved promising performance on sequential diagnosis prediction task, they are still facing two major challenges, such as, how to effectively model  irregular and temporal properties in EHR data and data insufficiency in healthcare information systems. In this paper, we propose an end-to-end transformer-based model, SETOR, to integrate medical ontology with visit information to mitigate the problem of data insufficiency, and to utilize neural ODE representations to learn hidden states for irregular intervals and visit discharges to effectively capture the irregular and temporal dependencies in EHR data. Although the proposed approach is focused on Electronic Health Record in healthcare domain, the core modules in this paper, e.g., ODE representations and ontological encoding, can be easily adapted into many other domains, such as, irregularly-sampled time series. An experiment is conducted to show that SETOR outperforms baselines with both sufficient and insufficient data. The representations of medical codes are visualized to illustrate the interpretability of the proposed model. The experimental results on the two real-world medical datasets demonstrate the effectiveness, robustness, and interpretability of the proposed model.

 \section*{Acknowledgment}
This work was supported in part by the Australian Research Council (ARC) under Grant LP180100654 and DE190100626.
\bibliographystyle{IEEEtran}
\bibliography{References.bib}

\end{document}